# Using Hankel Matrices for Dynamics-based Facial Emotion Recognition and Pain Detection (Pre-Print of Proceedings of AMFG CVPRW 2015)

Liliana Lo Presti and Marco La Cascia
DICGIM - University of Palermo
V.le delle Scienze, Ed. 6, 90128 Palermo (Italy)
liliana.lopresti@unipa.it

## Abstract

*This paper proposes a new approach to model the temporal dynamics of a sequence of facial expressions. To this purpose, a sequence of Face Image Descriptors (FID) is regarded as the output of a Linear Time Invariant (LTI) system. The temporal dynamics of such sequence of descriptors are represented by means of a Hankel matrix.*

*The paper presents different strategies to compute dynamics-based representation of a sequence of FID, and reports classification accuracy values of the proposed representations within different standard classification frameworks. The representations have been validated in two very challenging application domains: emotion recognition and pain detection. Experiments on two publicly available benchmarks and comparison with state-of-the-art approaches demonstrate that the dynamics-based FID representation attains competitive performance when off-the-shelf classification tools are adopted.*

## 1. Introduction

Facial expression analysis and emotion recognition are of interest in several domains such as human-computer interaction and social behavior understanding. Especially in socially assistive robotics [23] and computational behavioral science [24], [19], recognition of face expressions and emotions may help either to improve interactions with a robot, or to study people's social engagement in collaborative tasks [17], [35].

Moreover, face expression analysis could be useful in automatic pain monitoring, which in turn may help to ensure proper treatment to the patient [20]. Recent works, such as [21],[12], [27], [2],[25], [26], [9], [10] have focused on pain/no-pain detection and pain intensity estimation. These problems require the analysis of spontaneous facial expressions in the wild, namely under strong variations of head pose and face expressions. Pain detection suffers also of the difficulty to annotate the data in an objective way. Whilst patient's self-report is inexpensive and does not require for special skills, it has the drawback to be subjective, and it lacks of specific timing information [20]. Therefore, not only there are strong inter-patient variations in face expressions, but there are variations also in the pain self-reports.

Typically, approaches for emotion recognition tend to extract a representation of the face appearance, and adopt some classification framework. Some approaches [29] use a frame-based representation; others [6], [18], [30] use also temporal information. The latter works are motivated by the fact that face emotions are not instantaneous and the temporal evolution of face image descriptors (FIDs) can help to discriminate among emotions. Holistic representation of a sequence of FIDs, such as [29], lacks of temporal information, which instead proved to be useful [3].

In this paper, we propose to use the temporal dynamics of a sequence of FIDs to recognize among different emotions. Dynamics-based methods for emotion recognition have been proposed in [6] where a descriptor based on the movement of facial landmarks along the image sequence, and spatio-temporal appearance features are adopted. While [6] attempts to embed information about the dynamics at a feature representation level, works such as [18], [30] attempt to account for the temporal structure of the sequences of FIDs in the emotion model.

In contrast to these works, we propose to model a sequence of FIDs as the output of a Linear Time Invariant (LTI) system in order to perform dynamics-based emotion recognition. System identification [13], [8] or compressive sensing-based techniques [28] could be used to compare different emotion instances. However, previous works [14], [16] have shown that it may be possible to avoid the burden of performing system identification by representing the output of a LTI system through the corresponding Hankel matrix. Therefore, in this paper we explore the use of Hankel matrices in the domain of face analysis. The use of Hankel matrices, jointly with the dissimilarity score in [14], presents advantages in terms of space and time complexity.

In this paper, we propose different strategies to compute a dynamics-based representation that employs Hankel matrices. Our dynamics-based representation permits to easily compare sequences of different lengths. Comparison of sequences of different length is one of the major challenge in emotion classification. We present experiments in two different kinds of applications: emotion recognition and pain detection, and we conduct an extensive validation on two publicly available benchmarks. The first benchmark is the extended Cohn-Kanade dataset [19], which allows us to study the validity of this kind of feature representation for emotion recognition; the second dataset is the very challenging PAINFUL dataset [20], which allows us to study the proposed feature representation for pain localization in the wild. Our experiments show that, with standard and widely used classifiers such as nearest neighbor, linear SVM and HMM, our dynamics-based emotion representation allows us to consistently achieve state-of-the-art performance in emotion recognition in comparison to methods that use more complex machinery and costly training procedure.

The plan of the work is as follows. In Section 2, we present works that are related to our feature representation. In Section 3 we describe how we use a Hankel matrix to represent face emotions. In Section 4 we provide details of the adopted classification frameworks and in Section 5, we present extensive validation of the dynamics-based representation. Finally, in Section 6, we present conclusions and future directions.

## 2. Related Work

There is an extensive literature on face recognition [37], [15] and on facial expression analysis [35], [7]. Here we focus on works that try to embed the temporal structure of the sequence of facial expression either in the feature representation step or in the emotion-model.

In particular, [11] uses a Constrained Local Model (CLM) to obtain facial landmarks. Then it extracts patches around these markers. A sparse representation of the patches is obtained by applying non-negative matrix factorization. Classification is performed by least-square SVM. In [6], a descriptor based on the movement of facial landmark points over time, jointly with spatio-temporal appearance features is extracted for each face image sequence. The method attempts to measure horizontal and vertical movements of tracked landmarks of different face parts such as eyebrows, eyelids, cheeks, and lip corners. To account for temporal changes in the face appearance, Complete Local Binary Patterns from Three Orthogonal Planes (LBP-TOP) [36] are used. Classification is performed by SVM. In [22], restricted Boltzmann machine with local interactions (LRBM) is used to capture spatio-temporal patterns in the data. RBM is used as a generative model for data representation, and data need to be pre-aligned.

Since a sequence of FIDs is a time series, and may be affected by temporal warping, in [18] time-series kernel methods are used for emotional expression estimation using landmark data only. The work shows that emotion recognition may be done by adopting either the Dynamic Time Warping (DTW) kernel or the Global Alignment (GA) kernel [4, 5]. Our approach does not require any alignment of the data, and enables the comparison of sequences of different temporal duration.

To capture temporal information about the sequence of FIDs, Bayesian networks can be adopted. Wang et al. [33] propose to use Interval Temporal Bayesian Network (ITBN) to capture the spatial and temporal relations among primitive facial events. First, primitive facial events are identified, then ITBN is applied to model the interactions of primitives for expression recognition. In [30], a Bayesian approach is used to model dynamic facial expression temporal transitions. A face appearance representation is computed in terms of Local Binary Patterns (LBP), and an expression manifold is derived for multiple subjects. A Bayesian temporal model (similar to HMM with a non parametric observation model) of the manifold is used to represent facial expression dynamics.

In this paper, we model a sequence of FIDs by means of a dynamical system of unknown parameters. The parameters of the dynamical system are embedded in our Hankel matrix-based representation. Such representation aims at capturing the correlation among different face parts over time. Hankel matrices have already been adopted for action recognition in [14], which adopts a Hankel matrix-based bag-of-words approach, and in [16], which models an action as a sequence of Hankel matrices and uses a set of HMM trained in a discriminative way to model the switching between LTI systems. In contrast to these papers, we use Hankel matrices to describe the temporal dynamics of sequences of FIDs. We compare different strategies to compute a dynamics-based representation that employs Hankel matrices. We also test the effectiveness of our representation for emotion recognition within several standard classification frameworks.

## 3. Dynamics-based Emotion Representation

In this paper, a sequence of face images is processed to extract a feature representation on a frame-by-frame basis. This process yields to a time series of feature vectors $[y_o, \ldots, y_\tau]$, where $y_t$ is the feature representation associated with the $t$-th face image. Such temporal sequence can be regarded as the output of an LTI system of unknown pa-rameters [31].

## 3.1. Representation of Temporal Dynamics

In a linear time invariant system, two linear equations regulate the behavior of the system as follows:

$$\begin{aligned} x_{k+1} &= A \cdot x_k + w_k; \\ y_k &= C \cdot x_k. \end{aligned} \quad (1)$$

The first equation is known as the *state equation* and involves the variable $x_k \in R^u$, which represents the $u$-dimensional internal state of the LTI system. The second equation is known as the *measurement equation* and provides a link between the state of the system $x_k$ and the $v$-dimensional observable measurement $y_k$. In such equations the matrices A and C are constant over time, and $w_k \sim N(0, Q)$ is uncorrelated zero mean Gaussian measurement noise.

It is well known [32] that, given a sequence of output measurements $[y_o, \ldots, y_\tau]$ from Eq.1, its associated truncated block-Hankel matrix is

$$\widetilde{H} = \begin{bmatrix} y_0, & y_1, & y_2, & \ldots, & y_m \\ y_1, & y_2, & y_3, & \ldots, & y_{m+1} \\ \ldots & \ldots & \ldots & \ldots & \ldots \\ y_n, & y_{n+1}, & y_{n+2}, & \ldots, & y_\tau \end{bmatrix}, \quad (2)$$

where $n$ is the maximal order of the system, $\tau$ is the temporal length of the sequence, and it holds that $\tau = n + m - 1$.

The Hankel matrix embeds the observability matrix $\Gamma$ of the system, since $\widetilde{H} = \Gamma \cdot X$, where $X = [x_0, x_1, \cdots, x_\tau]$ is a matrix formed by the sequence of internal states of the LTI system.

As previously done in [14], [16], we normalize the Hankel matrix $\widetilde{H}$ as follows:

$$H = \frac{\widetilde{H}}{\sqrt{||\widetilde{H} \cdot \widetilde{H}^T||_F}}. \quad (3)$$

and compare two Hankel matrices $H_p$ and $H_q$ by:

$$d(H_p, H_q) = 2 - ||H_p \cdot H_p^T + H_q \cdot H_q^T||_F. \quad (4)$$

Such score, introduced in [14], does not define a distance. Instead, it roughly approximates the subspace angle between the spaces spanned by the columns of the Hankel matrices.

## 3.2. Dynamics-based Expression Representation

In this paper we propose to use a Hankel matrix to represent the dynamics of a sequence of face images whose feature representation yields to a time series of vectors $Y = [y_o, \ldots, y_\tau]$.

We compare three different dynamics-based emotion representations, that we describe in the following.

- **Single Hankel matrix representation:** this representation uses the whole time series $Y$ to build the Hankel matrix. We note that, even if the sequences may have different length, the matrix $H \cdot H^T$ used in Eq. 4 is a squared symmetric matrix and Hankel matrices of sequences of different lengths are easily comparable.

- **Sliding window-based representation:** while the former representation assumes segmentation of the frame sequence into emotions, this representation could overcome this requirement by representing $Y$ through a sequence of overlapping temporal window (similar to [16]). However, it may also limit the applicability of some classification frameworks (such as linear SVM) because sequence representations may have different lengths.

- **LTI Codebook-based representation:** in this representation, a bag-of-LTI-systems approach is used. First, the sequence $Y$ is represented by means of a single Hankel matrix $H$. From a training set, a codebook of LTI systems $C = \{C_i\}$, with $C_i$ representing a Hankel matrix, is computed by using K-medoids on the dissimilarity score in Eq. 4. The representation of a time series $Y$ is formed by concatenating the dissimilarity score of the Hankel matrix $H$ and each of the elements $C_i$ in the codebook. L2-normalization is applied on the extracted descriptor. Sequences of FIDs are represented by descriptors of the same length.

## 4. Adopted Classification Framework

We have tested our dynamics-based emotion representations within several classification frameworks in order to test their robustness.

- **Nearest Neighbor Classifier (NN):** given the dynamics-based representation of a test sequence, the predicted class is determined by the class label of the nearest sequence in the training set;

- **Codebook-based Support Vector Machine (CSVM):** Linear one-vs-all SVM models[1] are trained on the LTI codebook-based representation, and the estimated margin is used to classify the test sequence.

- **Dynamic Time Warping and NN (DTW+NN):** this method is applied to the sliding window-based representation. DTW[2] is used to align sequences of Hankel matrices. The Hankel matrices of each temporal window are compared through the score in Eq. 4. After aligning a test sequence with each sample in the training set, NN classifier is used to predict the test sequence class.

---
[1] We have used the Matlab implementation for linear SVM.
[2] We have used a slightly modified implementation of the code available at http://www.ee.columbia.edu/ln/rosa/matlab/dtw/

- **Nearest Neighbor and Majority Vote (NN+V):** this method assumes that the sliding window-based Hankel matrix representation is used. Inspired by [34], we use the NN classifier to predict the class label of each temporal window. Majority vote is used to predict the class of the test sequence.

- **Hidden Markov Model:** this method assumes the sliding window-based Hankel matrix representation is used. Similarly to [16], a HMM is used to model the transition from a LTI system to the other. In contrast to [16], we use standard HMM[3] with independently estimated state spaces. The number of states has been empirically set to 10. States are initialized by K-medoids and are not updated during the training procedure. We train an HMM for each class and use maximum-likelihood to classify a test sequence.

## 5. Experimental Results

This paper focuses on the analysis of face expressions in two challenging application domains: emotion recognition and pain detection. In the following, first we detail the frame-based representation adopted to obtain the measurements $[y_0, \cdots y_\tau]$, later we provide a brief description of each application domain and present our results.

### 5.1. Feature Extraction

Our formulation is general and can be adopted with several kinds of facial features. In this paper, to demonstrate the whole framework, we consider shape features provided by an active appearance model [19], [20]. Therefore, face expressions are represented as trajectories of 2D facial landmarks as shown in Fig. 1. To build the Hankel matrices, we use the following frame-based feature representations:

- concatenated 2D facial landmark coordinates (L);
- pairwise landmark distances (D);
- concatenation of pairwise landmark distances and landmark coordinates (L+D).

For each of these representations, principal component analysis (PCA) has been applied for noise and dimensionality reduction. We have selected a number of projections covering 99% of the total variance. The retained PCA coefficients are then used to build the dynamics-based representation as explained in Section 3.

---

[3]We used the HMM toolbox available at http://www.cs.ubc.ca/∼murphyk/Software/HMM/hmm.html. We modified the code in such a way that the observation model is an exponential distribution and each state is a LTI system represented by an exemplar Hankel matrix.

The adopted features are meant to represent the behavior of different parts of the face. The distance-based expression representation captures also the reciprocal relations among face parts (for example the joint movement of eyebrows and lips), and it is independent on the head movements on the image plane. The concatenation of distances and landmark coordinates permits to represent reciprocal relations of face parts given the face shape.

### 5.2. Emotion Recognition

Emotion recognition deals with the problem of inferring the emotion (such as fear, anger, surprise, etc.) given a sequence of face images. The main difficulty in this domain arises from the strong inter-subject variations, especially in some kind of emotions (such as sadness). Other challenges are connected with the difficulty to extract reliable feature representations due to illumination changes, biometric differences, head pose changes. Moreover, the lack of depth information makes emotion recognition more difficult due to ambiguities in the facial shape representation. To demonstrate the idea behind this paper, we restrict the attention to segmented emotion recognition in frontal view as done also in previous works such as [18], [1], [22], [33].

#### 5.2.1 Data and Validation Protocols

We have performed experiments for emotion recognition on the widely adopted Extended Cohn-Kanade dataset (CK+) [19]. This dataset provides facial expressions of 210 adults. Participants were instructed to perform several facial display representing either single or combinations of action units. Based on the coded action units and by means of a validation procedure of the assigned label, the segmented recording of the participants' emotions were classified into 7 categories: *angry, contempt, disgust, fear, happy, sadness, surprise*. In total there are 327 sequences of the 7 annotated emotions, performed by 118 different individuals. The number of frames of these sequences ranges in $[6, 71]$ with an average value of about $18 \pm 8.6$. The dataset provides landmark tracking results obtained by an active appearance model, which we use in our experiments. We adopted the validation protocol suggested in [19], which is leave-one-subject-out cross-validation.

#### 5.2.2 Emotion Recognition – Results

We have performed an extensive validation of the dynamics-based emotion representations whose results are reported in Table 1. The table reports the per-class classification accuracy values for each emotion class and the average accuracy value.

The table is divided in 4 parts. The first part compares dynamics-based representations when the Hankel matrix is computed over the whole sequence. We compare the single Hankel matrix representation with the nearest neighbor classifier against the LTI system codebook-based representation and linear one-vs-all SVMs.

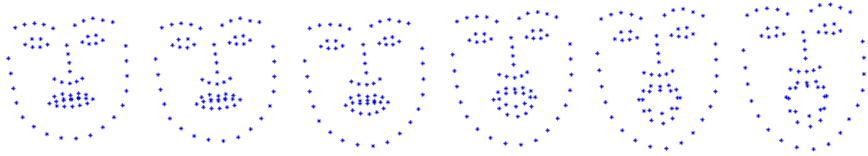

Figure 1. Sequence of 2D landmarks of six facial expressions (corresponding to surprise) from the CK+ dataset.

| Emotions: | Angry | Contempt | Disgust | Fear | Happy | Sadness | Surprise | Average |
|---|---|---|---|---|---|---|---|---|
| Hankel(L)+NN | 82.2 | 77.8 | 94.9 | 80 | **100** | 64.3 | 97.6 | 85.3 |
| Hankel(D)+NN | 88.9 | **83.3** | 96.6 | 84 | **100** | 67.9 | 98.8 | 88.5 |
| Hankel(L+D)+NN | 91.1 | **83.3** | 94.9 | 84 | **100** | 71.4 | 98.8 | **89.1** |
| Hankel(L)+CSVM | 86 | 75 | 92 | 85.6 | 98.3 | 74.3 | 95.9 | 86.7 |
| Hankel(D)+CSVM | 89.1 | 72.8 | 92.4 | **89.6** | 97 | 80.7 | 97.2 | 88.4 |
| Hankel(L+D)+CSVM | 89.8 | 73.9 | 90.8 | 89.2 | 97.4 | **81.8** | 97.7 | 88.7 |
| Hankel(L)+DTW+NN | 77.8 | 77.8 | 96.6 | 76 | **100** | 50 | 98.8 | 82.4 |
| Hankel(D)+DTW+NN | 82.2 | **83.3** | **98.3** | 72 | **100** | 60.7 | 98.8 | 85 |
| Hankel(L+D)+DTW+NN | 82.2 | **83.3** | **98.3** | 68 | **100** | 60.7 | 98.8 | 84.5 |
| Hankel(L)+NN+V | 86.7 | 77.8 | 91.5 | 84 | **100** | 75 | 98.8 | 87.7 |
| Hankel(D)+NN+V | 88.9 | **83.3** | 94.9 | 76 | **100** | 78.6 | 98.8 | 88.6 |
| Hankel(L+D)+NN+V | 91.1 | **83.3** | 94.9 | 76 | **100** | 78.6 | 98.8 | 89 |
| Hankel(L)+HMM | 84.4 | 72.2 | 89.8 | 88 | 95.6 | 64.3 | 95.2 | 84.2 |
| Hankel(D)+HMM | 82.2 | 66.7 | 93.2 | 80 | 97.1 | 53.6 | 97.6 | 81.5 |
| Hankel(L+D)+HMM | 84.4 | 66.7 | 93.2 | 80 | 95.6 | 57.1 | 97.6 | 82.1 |
| L+DTW+NN | 42.2 | 38.9 | 57.6 | 20 | 85.5 | 17.9 | 90.4 | 50.3 |
| D+DTW+NN | 57.8 | 44.4 | 57.6 | 20 | 91.3 | 28.6 | 95.2 | 56.4 |
| (L+D)+DTW+NN | 60 | 66.7 | 59.3 | 20 | 95.6 | 28.6 | 92.8 | 60.4 |
| L+NN+V | 33.3 | 44.4 | 42.4 | 36 | 82.6 | 14.3 | 83.1 | 48 |
| D+NN+V | 46.7 | 38.9 | 52.5 | 36 | 88.4 | 10.7 | 84.3 | 51.1 |
| (L+D)+NN+V | 42.2 | 50 | 54.2 | 28 | 88.4 | 25 | 89.2 | 53.9 |
| L+HMM | 20 | 33.3 | 45.8 | 32 | 66.7 | 14.3 | 85.5 | 42.5 |
| D+HMM | 44.4 | 38.9 | 57.6 | 48 | 82.6 | 39.3 | 84.3 | 56.5 |
| (L+D)+HMM | 48.9 | 50 | 44.1 | 40 | 81.2 | 21.4 | 84.4 | 52.8 |
| CK+ [19] | 35 | 25 | 68.4 | 21.7 | 98.4 | 4 | **100** | 50.4 |
| CLM-based [1] | 70.1 | 52.4 | 92.5 | 72.1 | 94.2 | 45.9 | 93.6 | 74.4 |
| LRBM [22] | **97.8** | 72.2 | 89.8 | 84 | **100** | 78.6 | 97.6 | 88.6 |
| ITBN [33] | 91.1 | 78.6 | 94 | 83.3 | 89.8 | 76 | 91.3 | 86.3 |

Table 1. Accuracy values (in %) for the Emotion Recognition task on the CK+ dataset. The last four rows report accuracy values of methods at the state-of-the-art on equal terms of features in input (namely, all the methods are landmark-based approaches). The adopted validation protocol is leave-one-subject-out cross-validation.

The second part of the table compares the sliding window-based representation within three classification frameworks: dynamic time warping and NN classifier, NN classifier and majority vote, hidden Markov models.

The third part of Table 1 reports the results obtained directly on the raw features (without computing any Hankel matrix) in order to highlight the advantage of using the dynamics-based representation. As the sequences have different lengths, we cannot apply the NN classifier directly on the raw features, but we are forced to align the sequences via DTW. We are not presenting results on the raw data via SVM because this experiment would be similar to the baseline method reported in [19] (in the lower part of the table). The fourth part of the table reports accuracy values of works at the state-of-the-art on equal terms of input data, which means that all the works we compare with use only

2D facial landmarks.

Due to the randomness in the codebook generation of the CSVM method and in the state initialization procedure in HMM, the experiments for CSVM and HMM have been repeated 10 times, and average accuracy values are reported.

Overall, the experimental results show that the dynamics-based emotion representation achieves state-of-the-art performance almost within all the tested classification frameworks. The highest accuracy values are obtained when the whole sequence is used. Among the sliding window-based approaches, only when adopting NN and majority vote the performance are comparable or higher than the one reported in [22]. These experiments suggest that probably emotions can be represented as the output of just one LTI system and there may be no dynamics switching as instead may happen in human actions [16]. The results also show how, in general, pairwise distances are more informative than 2D landmark trajectories. The concatenation of distances and landmarks (L+D) provides only a small improvement of the performance, with a general increase of the dimensionality of the representation. Finally, we note that the adoption of Hankel matrices permits to achieve an increase of more than 63% (on average) of the accuracy value obtained classifying directly the raw facial features.

## 5.3. Pain Detection

With respect to the former segmented emotion recognition task, pain detection is even more challenging due to the need of locating the pain event within the frame sequence. Pain may be a sporadic episode of varying duration, and painful facial expressions may vary greatly from subject-to-subject or be confused with other emotions.

In this paper we treat pain detection as a binary classification problem. We adopt a sliding window approach and classify each temporal window in order to detect the pain event. We empirically set the length of the temporal window to 10 frames.

### 5.3.1 Data and Validation Protocols

To test our dynamics-based FID representation for pain detection, we use the Painful dataset [20]. This dataset contains videos of patients' faces while they were moving their painful shoulder. The goal is that of recognizing between pain and no-pain events. The videos were annotated on a frame-per-frame basis with the Prkachin and Solomon pain intensity (PSPI) score [20], which ranges in [0, 15] where 0 means no pain, while a value greater than 0 indicates a certain intensity of pain. However, our method works on temporal windows and the validation requires a temporal window-based annotation. To account for this, we consider the integral score $IS$, obtained by summing the PSPI score in a sliding window. We set the label of the temporal window to 0 if $IS < \psi$, and to 1 if $IS \geq \psi$. Here $\psi$ indicates a threshold value on the integral score. A low value of $IS$ means that some frames in the temporal window have PSPI higher than 1; however, most of the frames in the window may score PSPI 0. To test the ability of our descriptor to represent pain events, we test our approach with different values of $\psi$ in leave-one-subject-out cross validation.

Finally, we note that this dataset is challenging also because faces are not always in frontal view, and the landmark points are affected by strong head movements. Therefore, on this dataset it is of particular interest the comparison between landmark-based and pairwise distance-based dynamics representation.

### 5.3.2 Pain Detection – Results

Table 2 reports the accuracy values of our sliding window-based Hankel matrix representation. We test such representation on landmarks (L) and on pairwise distance (D) measurements within two frameworks: NN and CSVM.

When adopting NN, the training set has been reduced by selecting only K medoids for each class, with K set to 300. When adopting CSVM, a codebook of 50 Hankel matrices is learned on the training set by K-medoids.

We report in columns the values of the confusion matrices for our binary classification experiment given the corresponding threshold value $\psi$. Positive indicates the pain event, while Negative indicates the no-pain event. Therefore TPR (true positive rate) and TNR (true negative rate) are the diagonal values of the confusion matrix. FNR (false negative rate) and FPR (false positive rate) are the extra-diagonal values. We also present the average classification accuracy (the average of the diagonal values).

Table 2 is divided into 4 parts. The first part reports values of the NN classifier when the Hankel matrix is computed directly on the landmark trajectories. The second part of the table reports values of the NN classifier when the Hankel matrix is computed on the pairwise landmark distances across time. By comparing the results, it is possible to observe that the pairwise distance-based representation achieves higher TPR for all the threshold values. The increase of the average accuracy for $\psi = 1$ is of around 18%, which is much higher than what observed on the CK+ dataset. We believe that, on these data, head motion might affect the landmark-based representation. On the contrary, pairwise distances are more invariant to head motion.

The third and fourth parts of the table reports accuracy values respectively for the landmark-based and pairwise distance-based measurements when adopting the LTI Codebook-based SVM approach. Again, the distance-based representation proves to be more discriminative than the landmark-based one. However, overall the accuracy values

| $\psi$: | $\geq 1$ | $\geq 11$ | $\geq 21$ | $\geq 31$ | $\geq 41$ |
|---|---|---|---|---|---|
| TPR (Hank(L)+NN) | 67.5 | 61 | 62.9 | 65 | 67.3 |
| FNR (Hank(L)+NN) | 32.5 | 39 | 37.1 | 35 | 32.7 |
| FPR (Hank(L)+NN) | 42.5 | 32 | 23.5 | 15.2 | 10 |
| TNR (Hank(L)+NN) | 57.5 | 68 | 76.5 | 84.8 | 90 |
| $\frac{TPR+TNR}{2}$ | 62.5 | 64.5 | 69.7 | 74.9 | 78.6 |
| TPR (Hank(D)+NN) | 75.8 | 76.9 | 78.7 | 80.7 | 80 |
| FNR (Hank(D)+NN) | 24.2 | 23.1 | 21.3 | 19.3 | 20 |
| FPR (Hank(D)+NN) | 27.6 | 22.2 | 16.8 | 11.4 | 7.8 |
| TNR (Hank(D)+NN) | 72.4 | 77.8 | 83.2 | 88.6 | 92.2 |
| $\frac{TPR+TNR}{2}$ | 74.1 | 77.3 | 80.9 | 84.6 | 86.1 |
| TPR (Hank(L)+SVM) | 57.2 | 54.3 | 49.4 | 45.3 | 54.4 |
| FNR (Hank(L)+SVM) | 42.8 | 45.7 | 50.6 | 54.3 | 45.6 |
| FPR (Hank(L)+SVM) | 44.9 | 43.7 | 33.2 | 29.2 | 28.7 |
| TNR (Hank(L)+SVM) | 55.1 | 56.3 | 66.8 | 70.8 | 71.3 |
| $\frac{TPR+TNR}{2}$ | 56.8 | 55.3 | 58.1 | 58.1 | 62.9 |
| TPR (Hank(D)+SVM) | 63.4 | 62.9 | 69 | 65.1 | 67.5 |
| FNR (Hank(D)+SVM) | 36.6 | 37.1 | 31 | 34.9 | 32.5 |
| FPR (Hank(D)+SVM) | 33 | 29.2 | 24.2 | 20.6 | 20.8 |
| TNR (Hank(D)+SVM) | 67 | 70.8 | 75.3 | 79.4 | 79.2 |
| $\frac{TPR+TNR}{2}$ | 65.2 | 66.9 | 72.4 | 72.2 | 73.4 |

Table 2. Accuracy of the pain detector at varying threshold values of the integral pain intensity score. TPR=True Positive Rate, FNR=False Negative Rate, FPR=False Positive Rate, True Negative Rate.

obtained with the adopted simple implementation of SVM are lower than the ones obtained via NN.

## 6. Conclusions and Future Work

In this paper we have proposed to adopt Hankel matrices to represent the dynamics of FIDs and recognize among different emotions. Whilst Hankel matrices have already been used in action recognition, at the best of our knowledge this paper is the first to adopt such kind of representation for face expression analysis. To study the performance of our dynamics-based emotion representations, we have performed extensive test within different standard classification frameworks on a widely used publicly available benchmark (CK+). Our experiments show that, on equal terms of classification framework, by using our dynamics-based emotion representations it is possible to achieve an increase of about 63% of the average accuracy values with respect of using directly the observed measurements. Overall, our approach achieves state-of-the-art performance by adopting standard classification frameworks.

We have also performed experiment on the Painful dataset to test if our representation can be used to detect pain events. The experiments show that the pairwise landmark distance-based dynamics representation is more invariant to head motion and, when using NN, it permits to obtain an increase of the average accuracy of about 18% with respect to the landmark-based representation.

The main limitation of our current approach is the need for reliable landmarks to represent face expressions. Despite the huge progress in this field, still facial landmark tracking is an open problem. On the other hand, our formulation is general and its not limited to 2D feature tracks. We therefore aim at extending our work considering appearance-based feature representation. Moreover, we have used a sliding window approach where windows have all the same temporal duration. We will explore the use of varying duration windows within a structure learning framework to automatically segment the emotion in a long sequence of face images.

## 7. Acknowledgement

This work was partially supported by the Italian MIUR grant PON01 01687, SINTESYS - Security and INTElligence SYStem.

## References

[1] S. W. Chew, P. Lucey, S. Lucey, J. Saragih, J. F. Cohn, and S. Sridharan. Person-independent facial expression detection using constrained local models. In *Proc. of Conf. and Workshop on Automatic Face & Gesture Recognition (FG)*, pages 915–920. IEEE, 2011.

[2] S. W. Chew, S. Lucey, P. Lucey, S. Sridharan, and J. F. Conn. Improved facial expression recognition via uni-hyperplane classification. In *Proc. of Conf. on Computer Vision and Pattern Recognition (CVPR)*, pages 2554–2561. IEEE, 2012.

[3] S. W. Chew, R. Rana, P. Lucey, S. Lucey, and S. Sridharan. Sparse temporal representations for facial expression recognition. In *Advances in Image and Video Technology*, pages 311–322. Springer, 2012.

[4] M. Cuturi. Fast global alignment kernels. In *Int. Conf. on Machine Learning (ICML)*, pages 929–936, 2011.

[5] M. Cuturi, J. Vert, O. Birkenes, and T. Matsui. A kernel for time series based on global alignments. In *Proc. of Int. Conf. on Acoustics, Speech and Signal Processing (ICASSP)*, volume 2, pages II–413. IEEE, 2007.

[6] H. Dibeklioğlu. *Enabling dynamics in face analysis*. PhD thesis, University of Amsterdam, 2014.

[7] B. Fasel and J. Luettin. Automatic facial expression analysis: a survey. *Pattern recognition*, 36(1):259–275, 2003.

[8] M. Fazel. *Matrix rank minimization with applications*. PhD thesis, PhD thesis, Stanford University, 2002.

[9] Z. Hammal and J. F. Cohn. Automatic detection of pain intensity. In *Proceedings of the 14th ACM international conference on Multimodal interaction*, pages 47–52. ACM, 2012.

[10] Z. Hammal and M. Kunz. Pain monitoring: A dynamic and context-sensitive system. *Pattern Recognition*, 45(4):1265–1280, 2012.

[11] L. A. Jeni, J. M. Girard, J. F. Cohn, and F. De La Torre. Continuous AU intensity estimation using localized, sparse

facial feature space. In *Proc. of Conf. on Automatic Face & Gesture Recognition (FG)*, pages 1–7. IEEE, 2013.

[12] S. Kaltwang, O. Rudovic, and M. Pantic. Continuous pain intensity estimation from facial expressions. In *Advances in Visual Computing*, pages 368–377. Springer, 2012.

[13] J. Klamka. Controllability of dynamical systems. a survey. *Bulletin of the Polish Academy of Sciences: Technical Sciences*, 61(2):335–342, 2013.

[14] B. Li, O. I. Camps, and M. Sznaier. Cross-view activity recognition using Hankelets. In *Proc. of Conf. on Computer Vision and Pattern Recognition (CVPR)*, pages 1362–1369. IEEE, 2012.

[15] L. Lo Presti and M. La Cascia. An on-line learning method for face association in personal photo collection. *Image and Vision Computing*, 30(4):306–316, 2012.

[16] L. Lo Presti, M. La Cascia, S. Sclaroff, and O. Camps. Gesture modeling by Hanklet-based hidden Markov model. In *Computer Vision – ACCV 2014*, volume 9005 of *Lecture Notes in Computer Science*, pages 529–546. Springer, 2015.

[17] L. Lo Presti, S. Sclaroff, and A. Rozga. Joint alignment and modeling of correlated behavior streams. In *Int. Conf. on Computer Vision-Workshops*, pages 730–737, 2013.

[18] A. Lorincz, L. A. Jeni, Z. Szabó, J. F. Cohn, and T. Kanade. Emotional expression classification using time-series kernels. In *Conf. on Computer Vision and Pattern Recognition Workshops (CVPRW)*, pages 889–895. IEEE, 2013.

[19] P. Lucey, J. F. Cohn, T. Kanade, J. Saragih, Z. Ambadar, and I. Matthews. The Extended Cohn-Kanade dataset (CK+): A complete dataset for action unit and emotion-specified expression. In *Proc. of Conf. on Computer Vision and Pattern Recognition Workshops (CVPRW)*, pages 94–101. IEEE, 2010.

[20] P. Lucey, J. F. Cohn, K. M. Prkachin, P. E. Solomon, and I. Matthews. Painful data: The UNBC-McMaster shoulder pain expression archive database. In *Proc. of Conf. and Workshop on Automatic Face & Gesture Recognition (FG)*, pages 57–64. IEEE, 2011.

[21] H. Meng and N. Bianchi-Berthouze. Affective state level recognition in naturalistic facial and vocal expressions. *IEEE Transactions on Cybernetics*, 44(3):315–328, 2014.

[22] S. Nie, Z. Wang, and Q. Ji. A generative restricted Boltzmann machine based method for high-dimensional motion data modeling. *Computer Vision and Image Understanding*, 2015.

[23] S. M. Rabbitt, A. E. Kazdin, and B. Scassellati. Integrating socially assistive robotics into mental healthcare interventions: Applications and recommendations for expanded use. *Clinical psychology review*, 35:35–46, 2015.

[24] J. M. Rehg et al. Decoding children's social behavior. In *Proc. of Conf. on Computer Vision and Pattern Recognition (CVPR)*, pages 3414–3421. IEEE, 2013.

[25] B. Romera-Paredes, A. Argyriou, N. Berthouze, and M. Pontil. Exploiting unrelated tasks in multi-task learning. In *International Conference on Artificial Intelligence and Statistics*, pages 951–959, 2012.

[26] B. Romera-Paredes, H. Aung, N. Bianchi-Berthouze, and M. Pontil. Multilinear multitask learning. In *Proceedings of the 30th International Conference on Machine Learning*, pages 1444–1452, 2013.

[27] O. Rudovic, V. Pavlovic, and M. Pantic. Automatic pain intensity estimation with heteroscedastic conditional ordinal random fields. In *Advances in Visual Computing*, pages 234–243. Springer, 2013.

[28] A. C. Sankaranarayanan, P. K. Turaga, R. G. Baraniuk, and R. Chellappa. Compressive acquisition of dynamic scenes. In *Proc. of European Conf. on Computer Vision (ECCV)*, pages 129–142. Springer, 2010.

[29] C. Shan, S. Gong, and P. W. McOwan. Robust facial expression recognition using local binary patterns. In *Int. Conf. on Image Processing*, volume 2, pages II–370. IEEE, 2005.

[30] C. Shan, S. Gong, and P. W. McOwan. Dynamic facial expression recognition using a Bayesian temporal manifold model. In *BMVC*, pages 297–306, 2006.

[31] E. D. Sontag. Nonlinear regulation: The piecewise linear approach. *IEEE Transactions on Automatic Control*, 26(2):346–358, 1981.

[32] M. Viberg. Subspace-based methods for the identification of linear time-invariant systems. *Automatica*, 31(12):1835–1851, 1995.

[33] Z. Wang, S. Wang, and Q. Ji. Capturing complex spatio-temporal relations among facial muscles for facial expression recognition. In *Conf. on Computer Vision and Pattern Recognition (CVPR)*, pages 3422–3429. IEEE, 2013.

[34] M. Zanfir, M. Leordeanu, and C. Sminchisescu. The moving pose: An efficient 3D kinematics descriptor for low-latency action recognition and detection. In *Proc. of Int. Conf. on Computer Vision (ICCV)*, pages 2752–2759. IEEE, 2013.

[35] Z. Zeng, M. Pantic, G. I. Roisman, and T. S. Huang. A survey of affect recognition methods: Audio, visual, and spontaneous expressions. *IEEE Trans. on Pattern Analysis and Machine Intelligence (PAMI)*, 31(1):39–58, 2009.

[36] G. Zhao and M. Pietikainen. Dynamic texture recognition using local binary patterns with an application to facial expressions. *IEEE Trans. on Pattern Analysis and Machine Intelligence (PAMI)*, 29(6):915–928, 2007.

[37] W. Zhao, R. Chellappa, P. J. Phillips, and A. Rosenfeld. Face recognition: A literature survey. *Acm Computing Surveys (CSUR)*, 35(4):399–458, 2003.